%% file: main.tex
\documentclass[10pt,twocolumn,letterpaper]{article}

\usepackage[pagenumbers]{cvpr}

\input{preamble}

\definecolor{linkblue}{rgb}{0.21,0.49,0.74}
\usepackage[pagebackref,breaklinks,colorlinks,allcolors=linkblue]{hyperref}
\hypersetup{
  pdftitle={MacroGS: Macro-Pixel Shared Gaussian Primitives for Arbitrary-Scale Light Field Super-Resolution},
  pdfauthor={Zeqiang Wei, Angulia Yang, Min Xu}
}

\title{MacroGS: Macro-Pixel Shared Gaussian Primitives for\\Arbitrary-Scale Light Field Super-Resolution}

\author{
Zeqiang Wei\textsuperscript{1}\qquad
Angulia Yang\textsuperscript{2}\qquad
Min Xu\textsuperscript{1,}\thanks{Corresponding author (email: xumin@cnu.edu.cn).}\\
\textsuperscript{1}Capital Normal University\qquad
\textsuperscript{2}National University of Singapore
}

\begin{document}
\maketitle
\input{sec/0_abstract}
\input{sec/1_intro}
\input{sec/2_related}
\input{sec/3_method}
\input{sec/4_experiments}
\input{sec/5_conclusion}

{
    \small
    \bibliographystyle{ieeenat_fullname}
    \bibliography{main}
}

\end{document}

%% file: preamble.tex
\usepackage{cuted}
\usepackage{currfile}
\usepackage{caption}

\usepackage{multirow}

\newcommand{\gain}[1]{\textsubscript{\textcolor{red}{\textbf{#1}}}}

\newcommand{\method}{MacroGS\xspace}

\graphicspath{{figures/}}
\newlength{\dispw} %

\newcommand{\bmu}{\boldsymbol{\mu}}
\newcommand{\bmub}{\bar{\boldsymbol{\mu}}}
\newcommand{\bd}{\mathbf{d}}
\newcommand{\bq}{\mathbf{q}}

\newcommand{\bz}{\mathbf{z}}
\newcommand{\bF}{\mathbf{F}}
\newcommand{\bG}{\mathbf{G}}
\newcommand{\be}{\mathbf{e}}
\newcommand{\bt}{\mathbf{t}}

\newcommand{\btau}{\boldsymbol{\tau}}
\newcommand{\bgamma}{\boldsymbol{\gamma}}
\newcommand{\bSigma}{\boldsymbol{\Sigma}}
\newcommand{\bphi}{\boldsymbol{\phi}}

%% file: sec/0_abstract.tex
\begin{abstract}
Arbitrary-scale light field spatial super-resolution reconstructs high-resolution views at any requested scale from low-resolution multi-view observations.
Gaussian primitives enable continuous reconstruction, but generating separate primitive sets for each view redundantly encodes shared scene structure across views and causes the primitive count grow linearly with the number of views.
To avoid such redundancy, we propose \method, a feed-forward decoder that jointly predicts a single shared Gaussian set per macro pixel rather than generating separate sets for each view. Disparity-dependent replacement and angle-conditioned appearance allow each set to be reused across all output views, keeping the predicted primitive count independent of the number of views at each requested scale.
The decoder iteratively predicts per-primitive disparities, retrieves the corresponding multi-view features, and updates the primitive states.
Gaussian splatting and residual refinement then reconstruct the output light field.
Despite being learned without geometric supervision, the predicted disparities strongly correlate with underlying scene geometry.
Experiments across five benchmarks demonstrate that \method integrates effectively with diverse light field encoders, improving reconstruction quality at fixed scales while achieving leading performance across most evaluated arbitrary-scale settings.
The decoder also achieves speedups of approximately \mbox{\textbf{\boldmath$2$--$10\times$}} over state-of-the-art implicit light field decoders and \mbox{\textbf{\boldmath$10$--$18\times$}} over the view-wise Gaussian baseline.
Our code is available at 
\href{https://github.com/Subury/MacroGS}{Github}.
\end{abstract}

%% file: sec/1_intro.tex
\section{Introduction}
\label{sec:intro}

\begin{figure}[t]
  \centering
  \includegraphics[width=\linewidth]{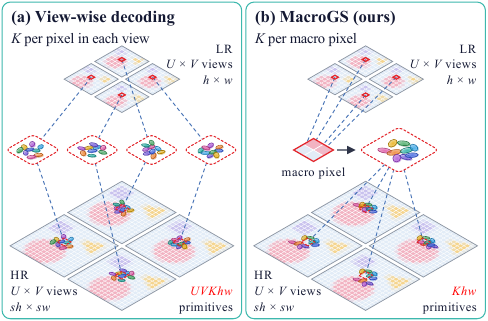}
  \caption{Gaussian representations of a $2\times2$ light field at one requested scale; colours mark the $K$ primitives. Only the primitives generated at the red pixel are drawn; every LR pixel generates its own set.
  (a)~Applying an image Gaussian decoder~\cite{gsasr,continuoussr,grape} separately to each SAI gives $UVKhw$ primitives ($K$ per pixel in each view) with separately predicted geometric parameters.
  (b)~\method jointly predicts $K$ primitives from the angular observations of each macro pixel and shares them across all SAIs, giving $Khw$ primitives. Each primitive's disparity controls its displacement across views.}
  \label{fig:teaser}
\end{figure}

Arbitrary-scale light field (LF) spatial super-resolution (LFSSR) reconstructs high-resolution sub-aperture images (SAIs) at a requested spatial scale from low-resolution multi-view observations, while retaining the input viewpoint grid.
The views observe common surfaces at displaced image locations~\cite{levoy1996}, providing complementary samples for recovering spatial detail~\cite{bishop2012}.
Recent encoders aggregate spatial and angular information into low-resolution features~\cite{distgssr,lft,epit,raslf}, which a decoder maps to high-resolution views.

Fixed-factor upsamplers support only predefined upsampling factors.
Coordinate-based implicit decoders allow arbitrary scales by predicting image values from queried coordinates and encoded features~\cite{liif,seiif,saeiif}.
However, reconstruction requires neural queries over the high-resolution grid of every view.
Explicit 2D Gaussian decoders instead predict primitives from image features and splat them at the requested resolution~\cite{gsasr,continuoussr,grape}.
Each primitive contributes to multiple output pixels, allowing its parameters to be reused across spatial samples.

For light fields, common scene structure also motivates sharing primitives across views.
A direct extension of a single-image Gaussian decoder applies it separately to each SAI, producing view-specific primitive sets (\cref{fig:teaser}a).
Even when the input features contain multi-view information, these sets do not share primitive identities or geometric parameters.
Common scene structure is therefore parameterized separately, and the primitive count grows with the number of views.

We introduce \method, a feed-forward decoder that jointly predicts Gaussian primitive sets and shares them across output views.
It changes the generation unit from an individual SAI pixel to a \emph{macro pixel}: the samples at one low-resolution spatial coordinate across the $U\times V$ views (\cref{fig:teaser}b).
For each requested scale, a macro pixel generates $K$ primitives that all output views reuse.
For input SAIs of spatial size $h\times w$, this gives $Khw$ primitives, compared with $UVKhw$ in a view-wise design with the same $K$ per location.
Feature reading and rendering still process all views.

Joint prediction must account for the fact that macro-pixel samples are not geometrically aligned and may observe different surfaces.
Assigning one disparity to the whole macro pixel would force its primitives to share the same cross-view displacement, including at foreground--background boundaries.
We instead predict a disparity for each primitive, allowing different displacements within the shared set.
Over a fixed number of rounds, the decoder predicts per-primitive disparities, reads features at the implied locations across views, and updates the corresponding primitive states.
Disparity thus guides both cross-view feature reading and primitive placement during rendering.

Primitive prediction and feature reading operate on the low-resolution grid.
Normalized Gaussian splatting renders the shared sets at the requested scale with angle-conditioned appearance, and a residual branch refines the output images.
Although the decoder is trained only for image reconstruction, the opacity-weighted means of its predicted disparities correlate strongly with scene geometry.

Across five benchmarks, \method achieves the highest average PSNR in eight of the ten evaluated encoder--scale settings, including out-of-scale factors beyond the training range.
Timed on the same GPU with the same input size, its decoder runs approximately $2$--$10\times$ as fast as SEIIF and SAEIIF~\cite{seiif,saeiif} and $10$--$18\times$ as fast as our view-wise GSASR adaptation~\cite{gsasr}.
Each encoder uses one arbitrary-scale model across all tested factors.
After scale-specific fine-tuning, \method improves all eight evaluated encoders over their original upsamplers at $\times2$ and $\times4$, with average PSNR gains of up to 0.70\,dB.

Our contributions are as follows.
\begin{itemize}
  \item A feed-forward macro-pixel Gaussian representation for arbitrary-scale light field spatial super-resolution that shares output primitives across views, reducing their count from $UVKhw$ to $Khw$ relative to a view-wise design with the same $K$.
  \item A per-primitive disparity parameterization that couples iterative cross-view feature reading with primitive displacement for rendering, allowing different cross-view trajectories within the shared set.
  \item Evaluation on five benchmarks demonstrating strong arbitrary-scale reconstruction and faster decoding, together with improvements across eight light field encoders after scale-specific fine-tuning.
\end{itemize}

%% file: sec/2_related.tex
\section{Related Work}
\label{sec:related}

\paragraph{Light field spatial super-resolution.}
LFSSR encoders aggregate the 4D signal through different feature organizations.
LF-InterNet~\cite{lfinternet} and DistgSSR~\cite{distgssr} use macro-pixel images, and LF-DFnet~\cite{lfdfnet} aligns views with deformable convolution.
LFT~\cite{lft}, EPIT~\cite{epit}, LF-DET~\cite{lfdet} and HLFSR~\cite{hlfsr} model spatial, angular and epipolar correlations with attention or hybrid designs~\cite{m2mt,oht}.
State-space methods process several LF subspaces~\cite{mlfsr,lfmamba,l2fmamba,lfmt}; RASLF~\cite{raslf} operates on sub-aperture, macro-pixel and epipolar representations.
Other methods use disparity explicitly, for reconstruction~\cite{bishop2012,wanner2014} or feature modulation~\cite{lfdgnet}.
Our focus is a shared output representation: \method jointly infers primitives from macro-pixel observations and reuses them across output views.

\paragraph{Arbitrary-scale upsampling.}
Meta-SR~\cite{metasr} predicts upsampling filters from the scale factor, and LIIF~\cite{liif} decodes the value at a continuous coordinate from nearby latent codes.
Subsequent implicit decoders, including LTE~\cite{lte}, CiaoSR~\cite{ciaosr}, DIIF~\cite{diif} and HIIF~\cite{hiif}, improve representation or query efficiency.
SEIIF~\cite{seiif} introduces epipolar decoding, selecting a sampling slope from predefined candidates by feature variance.
SAEIIF~\cite{saeiif} adds angular implicit functions on macro pixels for angular upsampling; for spatial super-resolution, where input and output share the same angular grid, its authors describe it as a deeper variant of SEIIF.
EASR~\cite{easr} and LF-ASSR~\cite{lfassr} address angular and joint spatial-angular upsampling, respectively.
\method predicts shared output primitives for spatial super-resolution, using continuous per-primitive disparities to guide cross-view inference.\looseness=-1

\paragraph{Gaussian primitives.}
3D Gaussian splatting~\cite{kerbl3dgs} represents scenes with anisotropic primitives and spherical-harmonic (SH) colour; GaussianImage~\cite{gaussianimage} fits image-plane 2D Gaussians using accumulation rasterization.
Feed-forward reconstruction methods include pixelSplat~\cite{pixelsplat} and MVSplat~\cite{mvsplat}.
GaussianSR~\cite{gaussiansr} places Gaussian fields at image pixels, whereas GSASR~\cite{gsasr} predicts movable primitives using learnable embeddings, attention and positional references.
ContinuousSR~\cite{continuoussr} and GRAPE~\cite{grape} further develop Gaussian image super-resolution.
For light fields, LFGaussian~\cite{lfgaussian} encodes cross-view features and estimates disparity before decoding separate Gaussian sets for source views.
It propagates and renders these sets, then fuses their features for angular super-resolution.
CVGS~\cite{cvgs} shares a central-view Gaussian set across views through predicted displacements, using per-scene fitting.
\method jointly predicts a shared set per macro pixel in a feed-forward decoder for spatial super-resolution, with per-primitive disparity and angle-conditioned appearance.

\paragraph{Geometric parameterization and iterative inference.}
N-dimensional splatting~\cite{ndgauss,gao6dgs,gao7dgs} models view- and time-dependent effects by slicing joint Gaussians over position and a query such as view direction, making each primitive's conditional mean and opacity query-dependent.
These representations are optimized per scene; concurrent work directly parameterizes the conditional mean instead of deriving it from the joint covariance~\cite{gaodgs}.
Layered depth images store multiple depth samples along viewing rays~\cite{ldi}; multiplane images~\cite{mpi,llff,deepview} and Fourier disparity layers~\cite{fdl,fdlsr} organize scene information over depth or disparity planes for view synthesis and reconstruction.
Our local primitives have individually predicted disparities, without assignment to common scene-wide planes.
Iterative estimation and sampling are established in RAFT~\cite{raft}, Deformable DETR~\cite{deformabledetr} and DeepView~\cite{deepview}.
Our loop applies this principle to each Gaussian's disparity: the current estimate specifies cross-view sampling positions, and sampled features update its state.
Slot Attention~\cite{slotattention} and GSASR~\cite{gsasr} also use multiple latent states; ours follow individual disparity-guided reading paths across views to infer a set shared by all views.

%% file: sec/3_method.tex
\section{Method}
\label{sec:method}

\begin{figure*}[t]
  \centering
  \includegraphics[width=\textwidth]{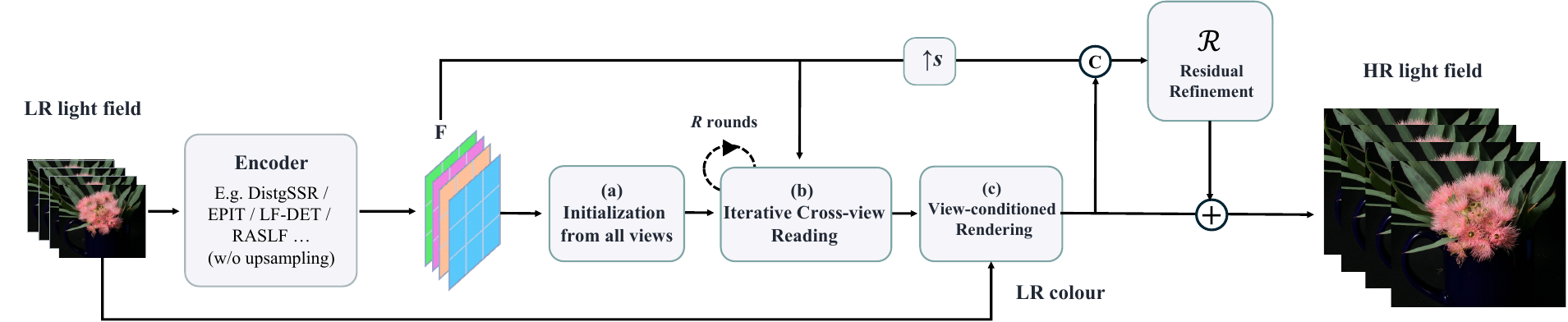}
  \caption{The framework of \method for arbitrary-scale LFSSR. From the encoder features $\bF$, \method infers one set $\mathcal G$ of $K$ Gaussian primitives per macro pixel, shared by all views, through (a) initialization from all views and (b) $R$ rounds of iterative cross-view reading. (c) View-conditioned rendering splats the set into every view at scale $s$, and a view-wise convolutional branch $\mathcal R$ adds a residual correction.}
  \label{fig:overview}
\end{figure*}

\subsection{Formulation and overview}
\label{sec:overview}

A light field with $U\times V$ views samples the scene from a regular grid of viewpoints. We index the views by $a\in\{1,\dots,UV\}$ and describe the position of view $a$ on the grid by its angular coordinates $\bq_a=(v_a,u_a)\in[-1,1]^2$, where $v_a$ and $u_a$ are its column and row indices normalized so that the central view lies at $(0,0)$ and the corner views at $(\pm1,\pm1)$.

Given an LR light field $L\in\mathbb{R}^{U\times V\times h\times w}$, arbitrary-scale LFSSR reconstructs the HR light field $\hat H\in\mathbb{R}^{U\times V\times sh\times sw}$ for a requested factor $s$, keeping the angular grid unchanged. As is common in LFSSR~\cite{ntire2023,seiif}, $L$ is the Y channel of the light field in YCbCr space, so all image values below are scalars.

We view an LFSSR network as the combination of an encoder and an upsampling module: the encoder extracts spatial-angular features $\bF\in\mathbb{R}^{U\times V\times h\times w\times C}$ from $L$, and \method serves as the upsampling module. For view $a$ and spatial position $\mathbf x$, we write $\bF(a,\mathbf x)\in\mathbb{R}^{C}$ for the feature and $L(a,\mathbf x)$ for the LR value; positions between pixel centres are handled by bilinear interpolation.

\method represents each macro pixel by a set of primitives shared across views (\cref{sec:rep}), infers the set through iterative cross-view reading (\cref{sec:reading}) and renders it into all HR views (\cref{sec:decoding}), as shown in \cref{fig:overview}.

\subsection{Macro-pixel primitive representation}
\label{sec:rep}

Let $\mathbf x_n$, $n=1,\ldots,hw$, be the LR pixel centres.
The pixels centred at $\mathbf x_n$ in all $UV$ views form macro pixel $n$.
For each macro pixel $n$, the decoder predicts a set of $K$ primitives, $\mathcal{G}_n=\{g_{n,k}\}_{k=1}^{K}$, where primitive $k$ has the parameters
\begin{equation*}
  g_{n,k}=\big(\bmub_{n,k},\ \bSigma_{n,k},\ \bd_{n,k},\ o_{n,k},\ \bgamma_{n,k}\big).
\end{equation*}
Here $\bmub_{n,k}\in\mathbb{R}^2$ and $\bSigma_{n,k}\in\mathbb{R}^{2\times2}$ are the centre and covariance in the central view, in LR pixel units. The disparity $\bd_{n,k}\in\mathbb{R}^2$ sets the horizontal and vertical displacement per unit angular coordinate. The appearance consists of a base opacity logit $o_{n,k}$ and SH coefficients $\bgamma_{n,k}=(\bgamma^{c}_{n,k},\bgamma^{\alpha}_{n,k})$ that adjust colour and opacity across views (\cref{sec:decoding}).
For each requested factor, all output views reuse the same $Khw$ primitives, whereas a per-view representation with the same $K$ would contain $UVKhw$. Feature reading and rendering still process all views.

\paragraph{Disparity-dependent placement.}
Each primitive moves linearly across views~\cite{wanner2014}; in view $a$ it is centred at
\begin{equation}
  \bmu_{n,k}^{(a)}=\bmub_{n,k}+\bd_{n,k}\odot\bq_a,
  \label{eq:displacement}
\end{equation}
where $\odot$ denotes element-wise multiplication. The covariance is shared across views.
Because each primitive has its own disparity, the primitives of one macro pixel can move at different rates, which provides geometric freedom near depth discontinuities. The same disparity also determines where the primitive reads features (\cref{sec:reading}) and samples the LR input (\cref{sec:decoding}).

\subsection{Disparity-guided primitive inference}
\label{sec:reading}

The $k$-th primitive of macro pixel $n$ is associated with a latent state $\bz_{n,k}\in\mathbb{R}^{C}$, a feature vector that starts from a learnable query $\be$ shared by all states, is refined over $R$ rounds and finally yields the attributes of the primitive.
Each state also carries a fixed template offset $\btau_k$, which anchors the centre of its primitive relative to $\mathbf x_n$, and an identity embedding $\bt_k=\mathrm{MLP}(\btau_k)$. The $K$ template offsets ($K=9$ by default) form a regular $\sqrt K\times\sqrt K$ lattice inside the LR pixel, so the primitives of a macro pixel start spread over its area. A scale embedding $\bphi(s)$ conditions the states on the requested factor.

\paragraph{Initialization from all views.}
Each state first attends to the $UV$ features of its macro pixel,
\begin{equation*}
  \bz_{n,k}^{(0)}=\mathrm{MHA}\big(\be+\bt_k,\ \{\bF(a,\mathbf{x}_n)+\boldsymbol\nu_a\}_{a=1}^{UV}\big)+\bt_k+\bphi(s),
\end{equation*}
where $\mathrm{MHA}(\mathbf q,\mathcal S)$ is multi-head attention with query $\mathbf q$ and keys and values $\mathcal S$; $\boldsymbol\nu_a$ is a learnable view embedding.

\paragraph{Iterative cross-view reading.}
Co-located features show the same scene point in all views only at zero disparity. Reading one point across views requires its disparity, which is in turn estimated from the features read.
Each round $r=1,\ldots,R$ ($R=3$ by default) therefore predicts a disparity from the current state, reads the feature $\bG^{(r)}_{n,k,a}$ of each view $a$ at the position it gives and updates the state,
\begin{equation}
\begin{aligned}
  &\bd_{n,k}^{(r)}=d_{\max}\tanh\big(\psi(\bz_{n,k}^{(r-1)})\big),\\
  &\bG_{n,k,a}^{(r)}=\bF\big(a,\ \mathbf{x}_n+\bd_{n,k}^{(r)}\odot\bq_a\big),\\
  &\bz_{n,k}^{(r)}=\mathrm{Blk}_r\big(\bz_{n,k}^{(r-1)}+\bt_k,\ \{\bG_{n,k,a}^{(r)}\}_{a=1}^{UV}\big)+\bphi(s),
\end{aligned}
\label{eq:reading}
\end{equation}
where $\psi$ is a two-layer MLP shared across rounds, the hyperparameter $d_{\max}$ bounds each disparity component, and each round has its own update block $\mathrm{Blk}_r$.
Each round thus predicts its disparity from the readings of the previous round and can correct their positions. Bilinear sampling allows reconstruction gradients to reach $\psi$.

After reading, a state has to summarize its readings, agree with the other primitives of its macro pixel, which share its output pixels, and use context from neighbouring macro pixels. With input $\mathbf h^0_{n,k}=\bz^{(r-1)}_{n,k}+\bt_k$, the four residual sub-layers of the update block $\mathrm{Blk}_r$ give
\begin{equation}
\begin{aligned}
  \mathbf h^{1}_{n,k}&=\mathbf h^{0}_{n,k} + \mathrm{MHA}\big(\mathbf h^{0}_{n,k},\ \{G^{(r)}_{n,k,a}+\boldsymbol\nu_a\}_{a=1}^{UV}\big),\\
  \mathbf h^{2}_{n,k}&=\mathbf h^{1}_{n,k}+\mathrm{MHA}\big(\mathbf h^{1}_{n,k},\ \{\mathbf h^{1}_{n,j}\}_{j=1}^{K}\big),\\
  \mathbf h^{3}_{n,k}&=\mathbf h^{2}_{n,k}+\mathrm{DWConv}\big(\{\mathbf h^{2}_{m,k}\}_{m\in\mathcal N(n)}\big),\\
  \mathbf h^{4}_{n,k}&=\mathbf h^{3}_{n,k}+\mathrm{FFN}\big(\mathbf h^{3}_{n,k}\big),
\end{aligned}
\label{eq:update}
\end{equation}
and the block returns $\mathbf h^4_{n,k}$.
\emph{Aggregate views} ($\mathbf h^1$) attends from each state to its own readings $\bG^{(r)}_{n,k,a}$, tagged with the view embeddings $\boldsymbol\nu_a$, and fuses the result with the state through an MLP on their concatenation $[\cdot,\cdot]$. Only the readings depend on the current disparity: at the right disparity they show one surface point in the views where it is visible, and per-view weights can favour consistent observations.
\emph{Exchange states} ($\mathbf h^2$) attends from each state to the $K$ states of its macro pixel.
\emph{Mix neighbours} ($\mathbf h^3$) applies $\mathrm{DWConv}$, a depthwise-separable convolution, over the $3\times3$ LR neighbourhood $\mathcal N(n)$ separately for each state index $k$, and \emph{transform states} ($\mathbf h^4$) is a per-state feed-forward network $\mathrm{FFN}$.

\paragraph{Final primitive attributes.}
After the final reading round, four lightweight MLP heads map each final state $\mathbf z=\bz^{(R)}_{n,k}$ to the attributes of its Gaussian primitive. We omit the indices $n,k$ below:
\begin{equation*}
\begin{aligned}
  (\boldsymbol\delta,\sigma,o)&=h_{\rm geo}(\mathbf z), & (\kappa,\rho)&=h_{\rm shape}(\mathbf z),\\
  \bgamma^{c}&=h_{c}(\mathbf z), & \bgamma^{\alpha}&=h_{\alpha}(\mathbf z).
\end{aligned}
\end{equation*}
The heads include the output transformations that constrain these attributes to valid ranges.
For position, the geometry head predicts a small offset $\boldsymbol\delta$ from the fixed template location, giving the centre in the central view as $\bmub=\mathbf x_n+\btau_k+\boldsymbol\delta$.
The offset is bounded to a fraction of an LR pixel, keeping the primitive near its template location.
Applying the reading-loop disparity head once more gives $\bd=d_{\max}\tanh\big(\psi(\mathbf z)\big)$, which determines how the centre moves across views.

The spatial footprint of a primitive can be understood as an ellipse.
The base scale $\sigma>0$ controls its overall size in LR pixels, $\kappa>0$ adjusts its relative horizontal and vertical spread, and $\rho\in(-1,1)$ allows the footprint to tilt.
Together, they define the covariance:
\begin{equation*}
  \bSigma=\sigma^2\begin{bmatrix}
  \kappa^2 & \rho\\
  \rho & \kappa^{-2}
  \end{bmatrix}.
\end{equation*}

For appearance, the geometry head also predicts a base opacity logit $o$.
The remaining two heads predict SH coefficients $\bgamma^c$ and $\bgamma^\alpha$, which adjust colour and opacity across views during rendering.

\subsection{View-conditioned rendering at arbitrary scale}
\label{sec:decoding}

Once the shared primitives have been inferred, \method renders each view at the requested resolution.
Rendering proceeds in three steps: determining the colour and opacity of each primitive for the view, blending their contributions on the HR grid, and refining the resulting image.

\paragraph{Angle-conditioned appearance.}
Each primitive obtains its base colour from the LR image of the view being rendered.
For view $a$, we sample the image at $\mathbf x_n$ displaced by the primitive's disparity:
\begin{equation*}
  \tilde L_{n,k}^{(a)}=L\big(a,\ \mathbf{x}_n+\bd_{n,k}\odot\bq_a\big).
\end{equation*}
The sampling position depends only on $\mathbf x_n$ and the disparity, while the rendering centre additionally includes the template and learned offsets defined in \cref{sec:reading}.
These offsets therefore adjust where the primitive is drawn without changing its colour sampling position.

The primitive's colour combines the sampled LR value with a view-dependent SH correction. The correction is added in the logit domain, with $\operatorname{logit}$ the inverse of the sigmoid, so a zero correction returns the sampled value and the colour stays in $(0,1)$. Shared SH coefficients also adjust its opacity across views.
We obtain the viewing direction from the angular coordinates of view $a$ and evaluate the real SH basis functions up to degree two to adjust the sampled colour and base opacity:
\begin{equation}
\begin{aligned}
  \boldsymbol\omega_a&=\frac{(v_a,u_a,1)}{\sqrt{1+u_a^2+v_a^2}},\\
  c_{n,k}^{(a)}&=\operatorname{sigmoid}\!\bigl(\operatorname{logit}(\tilde L_{n,k}^{(a)})\!+\!\textstyle\sum_m\gamma_{n,k}^{c,m}\mathrm{SH}_m(\boldsymbol\omega_a)\bigr),\\
  \alpha_{n,k}^{(a)}&=\operatorname{sigmoid}\!\bigl(o_{n,k}\!+\!\textstyle\sum_m\gamma_{n,k}^{\alpha,m}\mathrm{SH}_m(\boldsymbol\omega_a)\bigr).
\end{aligned}
\label{eq:appearance}
\end{equation}

\begin{table*}[t]
  \centering
  \caption{Fixed-scale comparison in PSNR (dB). Each encoder is evaluated with its original upsampler (baseline) and with \method; red subscripts indicate PSNR increases. Baselines are trained for each scale, while \method is fine-tuned from its arbitrary-scale model for each target scale following SEIIF~\cite{seiif}.}
  \label{tab:fixed}
  \footnotesize
  \setlength{\tabcolsep}{2.2pt}
  \begin{tabular}{l*{12}{l}}
    \toprule
    \multirow{2}{*}{Method} & \multicolumn{2}{c}{EPFL~\cite{epfl}} & \multicolumn{2}{c}{HCInew~\cite{hcinew}} & \multicolumn{2}{c}{HCIold~\cite{hciold}} & \multicolumn{2}{c}{INRIA~\cite{inria}} & \multicolumn{2}{c}{STFgantry~\cite{stfgantry}} & \multicolumn{2}{c}{Average} \\
    \cmidrule(lr){2-3}\cmidrule(lr){4-5}\cmidrule(lr){6-7}\cmidrule(lr){8-9}\cmidrule(lr){10-11}\cmidrule(lr){12-13}
    & \multicolumn{1}{c}{$\times2$} & \multicolumn{1}{c}{$\times4$} & \multicolumn{1}{c}{$\times2$} & \multicolumn{1}{c}{$\times4$} & \multicolumn{1}{c}{$\times2$} & \multicolumn{1}{c}{$\times4$} & \multicolumn{1}{c}{$\times2$} & \multicolumn{1}{c}{$\times4$} & \multicolumn{1}{c}{$\times2$} & \multicolumn{1}{c}{$\times4$} & \multicolumn{1}{c}{$\times2$} & \multicolumn{1}{c}{$\times4$} \\
    \midrule
    LF-InterNet~\cite{lfinternet} & 34.11 & 28.81 & 37.17 & 30.96 & 44.57 & 37.15 & 35.83 & 30.78 & 38.44 & 30.37 & 38.02 & 31.61 \\
    LF-InterNet + \method & 34.47\gain{0.36} & 29.36\gain{0.55} & 37.68\gain{0.51} & 31.62\gain{0.66} & 44.85\gain{0.28} & 37.75\gain{0.60} & 36.15\gain{0.32} & 31.27\gain{0.49} & 39.64\gain{1.20} & 31.56\gain{1.19} & 38.56\gain{0.54} & 32.31\gain{0.70} \\
    \midrule
    LF-DFnet~\cite{lfdfnet} & 34.40 & 28.85 & 37.60 & 31.29 & 44.36 & 37.37 & 36.34 & 30.85 & 39.94 & 31.26 & 38.53 & 31.92 \\
    LF-DFnet + \method & 34.71\gain{0.31} & 29.53\gain{0.68} & 38.03\gain{0.43} & 31.74\gain{0.45} & 44.90\gain{0.54} & 37.96\gain{0.59} & 36.65\gain{0.31} & 31.68\gain{0.83} & 40.61\gain{0.67} & 31.92\gain{0.66} & 38.98\gain{0.45} & 32.57\gain{0.65} \\
    \midrule
    LF-IINet~\cite{lfiinet} & 34.74 & 29.05 & 37.77 & 31.33 & 44.85 & 37.62 & 36.56 & 31.04 & 39.89 & 31.26 & 38.76 & 32.06 \\
    LF-IINet + \method & 34.81\gain{0.07} & 29.53\gain{0.48} & 37.96\gain{0.19} & 31.71\gain{0.38} & 44.96\gain{0.11} & 37.84\gain{0.22} & 36.57\gain{0.01} & 31.47\gain{0.43} & 40.38\gain{0.49} & 31.78\gain{0.52} & 38.94\gain{0.18} & 32.47\gain{0.41} \\
    \midrule
    LFT~\cite{lft} & 34.80 & 29.26 & 37.84 & 31.46 & 44.52 & 37.63 & 36.59 & 31.21 & 40.51 & 31.86 & 38.85 & 32.28 \\
    LFT + \method & 34.85\gain{0.05} & 29.66\gain{0.40} & 37.90\gain{0.06} & 31.67\gain{0.21} & 44.82\gain{0.30} & 37.83\gain{0.20} & 36.63\gain{0.04} & 31.64\gain{0.43} & 40.82\gain{0.31} & 32.28\gain{0.42} & 39.01\gain{0.16} & 32.62\gain{0.34} \\
    \midrule
    DistgSSR~\cite{distgssr} & 34.81 & 28.99 & 37.96 & 31.38 & 44.94 & 37.56 & 36.59 & 30.99 & 40.40 & 31.65 & 38.94 & 32.12 \\
    DistgSSR + \method & 35.05\gain{0.24} & 29.68\gain{0.69} & 38.20\gain{0.24} & 31.91\gain{0.53} & 45.07\gain{0.13} & 38.09\gain{0.53} & 36.79\gain{0.20} & 31.63\gain{0.64} & 40.98\gain{0.58} & 32.47\gain{0.82} & 39.22\gain{0.28} & 32.76\gain{0.64} \\
    \midrule
    EPIT~\cite{epit} & 34.83 & 29.34 & 38.23 & 31.51 & 45.07 & 37.68 & 36.67 & 31.37 & 42.17 & 32.18 & 39.39 & 32.42 \\
    EPIT + \method & 34.97\gain{0.14} & 29.90\gain{0.56} & 38.47\gain{0.24} & 31.89\gain{0.38} & 45.38\gain{0.31} & 38.07\gain{0.39} & 36.70\gain{0.03} & 31.98\gain{0.61} & 42.75\gain{0.58} & 32.74\gain{0.56} & 39.66\gain{0.27} & 32.92\gain{0.50} \\
    \midrule
    LF-DET~\cite{lfdet} & 35.26 & 29.47 & 38.31 & 31.56 & 44.99 & 37.84 & 36.95 & 31.39 & 41.76 & 32.14 & 39.45 & 32.48 \\
    LF-DET + \method & 35.35\gain{0.09} & 30.09\gain{0.62} & 38.52\gain{0.21} & 31.97\gain{0.41} & 45.13\gain{0.14} & 38.12\gain{0.28} & 36.99\gain{0.04} & 32.00\gain{0.61} & 42.14\gain{0.38} & 32.79\gain{0.65} & 39.62\gain{0.17} & 32.99\gain{0.51} \\
    \midrule
    RASLF~\cite{raslf} & 35.17 & 29.76 & 38.43 & 31.67 & 45.31 & 37.89 & 36.99 & 31.76 & 41.87 & 32.37 & 39.55 & 32.69 \\
    RASLF + \method & 35.30\gain{0.13} & 30.09\gain{0.33} & 38.52\gain{0.09} & 31.97\gain{0.30} & 45.48\gain{0.17} & 38.16\gain{0.27} & 37.05\gain{0.06} & 32.13\gain{0.37} & 41.89\gain{0.02} & 32.85\gain{0.48} & 39.65\gain{0.10} & 33.04\gain{0.35} \\
    \bottomrule
  \end{tabular}
\end{table*}

\paragraph{Rendering at the requested scale.}
Using continuous coordinates with the origin at the image corner, let $\mathbf S=\mathrm{diag}(s_x,s_y)$ contain the horizontal and vertical scale factors.
Each primitive spreads its colour over nearby HR pixels according to its scaled Gaussian footprint. At pixel $\mathbf p$, its spatial weight is
\begin{equation*}
\begin{aligned}
  w_{n,k}^{(a)}(\mathbf p)&=\exp\Big(-\tfrac{1}{2}\boldsymbol\Delta^{\top}\big(\mathbf S\bSigma_{n,k}\mathbf S^{\top}\big)^{-1}\boldsymbol\Delta\Big),\\
  \boldsymbol\Delta&=\mathbf p-\mathbf S\bmu_{n,k}^{(a)}.
\end{aligned}
\end{equation*}
Multiplying this weight by the view-dependent opacity determines the primitive's contribution.
We combine the contributions through normalized splatting:
\begin{equation}
  \hat I_a(\mathbf p)=\frac{\sum_{n,k}w_{n,k}^{(a)}(\mathbf p)\,\alpha_{n,k}^{(a)}c_{n,k}^{(a)}}{\max\big(\sum_{n,k}w_{n,k}^{(a)}(\mathbf p)\,\alpha_{n,k}^{(a)},\,\epsilon\big)}.
  \label{eq:splat}
\end{equation}
The denominator normalizes the accumulated weights, and $\epsilon$ prevents division by zero.
Gaussian weights are evaluated within a local window around each centre.
This renderer blends primitive colours without depth sorting.

\paragraph{Residual refinement.}
To refine details in the splatted image, a convolutional branch predicts an additive correction from $\hat I_a$ and the encoder features of the same view:
\begin{equation}
  \hat H_a=\hat I_a+\mathcal R\big([\bF^{\uparrow}_a,\hat I_a]\big).
  \label{eq:residual}
\end{equation}
Here, $\bF^{\uparrow}_a$ denotes the encoder feature map bilinearly upsampled to the target resolution, and $[\cdot,\cdot]$ denotes channel concatenation.
The branch $\mathcal R$ is applied separately to each view with shared weights.
Primitive inference and cross-view reading operate on the LR grid, while splatting and residual refinement produce the image on the HR grid.

%% file: sec/4_experiments.tex
\section{Experiments}
\label{sec:experiments}

\begin{table*}[t]
  \centering
  \caption{Arbitrary-scale comparison: decoder FLOPs (G) / decoder time (s) / PSNR (dB) averaged over the five benchmarks; parameters count the decoder. All times are for a $5\times5\times32\times32$ input on a Tesla V100. Arbitrary-scale methods use one model per encoder for all factors; baselines use scale-specific models. Best and second-best PSNR in each block are in bold and underlined. OOM: out of memory~\cite{seiif}.}
  \label{tab:arbitrary}
  \footnotesize
  \setlength{\tabcolsep}{2.6pt}
  \begin{tabular}{lc*{5}{r@{\,/\,}c@{\,/\,}l}}
    \toprule
    \multirow{2}{*}{Method} & \multirow{2}{*}{Params (M)} & \multicolumn{9}{c}{FLOPs (G) / Time (s) / PSNR (in-scale)} & \multicolumn{6}{c}{FLOPs (G) / Time (s) / PSNR (out-of-scale)} \\
    \cmidrule(lr){3-11}\cmidrule(lr){12-17}
    & & \multicolumn{3}{c}{$\times2$} & \multicolumn{3}{c}{$\times3$} & \multicolumn{3}{c}{$\times4$} & \multicolumn{3}{c}{$\times6$} & \multicolumn{3}{c}{$\times8$} \\
    \midrule
    DistgSSR~\cite{distgssr} (Baseline) & [0.02, 0.07] & 0.4 & 0.04 & 38.94 & 1.0 & 0.04 & 34.70 & 1.7 & 0.04 & 32.12 & \multicolumn{3}{c}{--} & \multicolumn{3}{c}{--} \\
    \multicolumn{17}{l}{\textit{Implicit neural fields}} \\
    DistgSSR + LIIF~\cite{liif} & 0.09 & 35.6 & 0.07 & 38.90 & 79.9 & 0.11 & 34.81 & 142.0 & 0.17 & 32.42 & 319.5 & 0.38 & 29.39 & 568.0 & 0.68 & 27.47 \\
    DistgSSR + LTE~\cite{lte} & 0.13 & 22.2 & 0.07 & 38.94 & 47.5 & 0.11 & 34.88 & 82.9 & 0.16 & 32.46 & 184.1 & 0.32 & 29.43 & 325.8 & 0.57 & 27.53 \\
    DistgSSR + DIIF~\cite{diif} & 0.12 & 20.4 & 0.06 & 38.92 & 33.1 & 0.06 & 34.84 & 47.5 & 0.07 & 32.44 & 81.4 & 0.11 & 29.41 & 122.1 & 0.14 & 27.54 \\
    DistgSSR + CiaoSR~\cite{ciaosr} & 0.22 & 57.0 & 0.11 & 39.00 & 127.8 & 0.17 & 34.92 & 226.9 & 0.26 & 32.50 & 510.1 & 0.52 & 29.50 & 906.5 & 0.98 & OOM \\
    DistgSSR + SEIIF~\cite{seiif} & 0.12 & 8.4 & 0.07 & 39.01 & 16.6 & 0.08 & 34.93 & 27.3 & 0.08 & 32.53 & 56.5 & 0.12 & 29.48 & 96.2 & 0.16 & 27.60 \\
    DistgSSR + SAEIIF~\cite{saeiif} & 0.26 & 15 & 0.11 & \underline{39.03} & 26 & 0.13 & \underline{34.96} & 41 & 0.16 & \underline{32.55} & 80 & 0.30 & \underline{29.53} & 132 & 0.48 & 27.63 \\
    \multicolumn{17}{l}{\textit{Gaussian splatting}} \\
    DistgSSR + GSASR~\cite{gsasr} & 2.57 & 193.1 & 0.30 & 38.91 & 199.5 & 0.36 & 34.86 & 208.6 & 0.43 & 32.50 & 233.9 & 0.63 & 29.47 & 269.8 & 0.92 & \underline{27.69} \\
    DistgSSR + \method (Ours) & 0.37 & 16.2 & 0.03 & \textbf{39.19} & 25.9 & 0.03 & \textbf{35.14} & 39.4 & 0.03 & \textbf{32.76} & 78.0 & 0.04 & \textbf{29.58} & 132.2 & 0.05 & \textbf{27.73} \\
    \midrule
    EPIT~\cite{epit} (Baseline) & [0.02, 0.07] & 0.5 & 0.05 & 39.39 & 1.1 & 0.05 & 34.98 & 1.9 & 0.05 & 32.42 & \multicolumn{3}{c}{--} & \multicolumn{3}{c}{--} \\
    \multicolumn{17}{l}{\textit{Implicit neural fields}} \\
    EPIT + LIIF~\cite{liif} & 0.09 & 35.6 & 0.07 & 39.50 & 79.9 & 0.11 & 35.13 & 142.0 & 0.17 & 32.63 & 319.5 & 0.39 & 29.53 & 568.0 & 0.68 & 27.63 \\
    EPIT + LTE~\cite{lte} & 0.13 & 22.2 & 0.08 & 39.53 & 47.5 & 0.11 & 35.15 & 82.9 & 0.17 & 32.63 & 184.1 & 0.32 & 29.52 & 325.8 & 0.57 & 27.62 \\
    EPIT + DIIF~\cite{diif} & 0.12 & 20.4 & 0.06 & 39.51 & 33.1 & 0.07 & 35.16 & 47.5 & 0.08 & 32.66 & 81.4 & 0.11 & 29.55 & 122.1 & 0.14 & 27.64 \\
    EPIT + CiaoSR~\cite{ciaosr} & 0.22 & 57.0 & 0.11 & \underline{39.59} & 127.8 & 0.17 & 35.21 & 226.9 & 0.26 & 32.72 & 510.1 & 0.52 & 29.61 & 906.5 & 0.99 & OOM \\
    EPIT + SEIIF~\cite{seiif} & 0.12 & 8.4 & 0.07 & \textbf{39.60} & 16.6 & 0.08 & 35.25 & 27.3 & 0.09 & 32.76 & 56.5 & 0.12 & 29.61 & 96.2 & 0.17 & 27.68 \\
    EPIT + SAEIIF~\cite{saeiif} & 0.26 & 15 & 0.11 & \textbf{39.60} & 26 & 0.13 & 35.27 & 41 & 0.17 & \underline{32.79} & 80 & 0.30 & \textbf{29.67} & 132 & 0.50 & \underline{27.75} \\
    \multicolumn{17}{l}{\textit{Gaussian splatting}} \\
    EPIT + GSASR~\cite{gsasr} & 2.57 & 193.1 & 0.30 & 39.31 & 199.5 & 0.36 & \underline{35.30} & 208.6 & 0.43 & 32.56 & 233.9 & 0.63 & 29.49 & 269.8 & 0.92 & 27.64 \\
    EPIT + \method (Ours) & 0.37 & 16.2 & 0.03 & 39.58 & 25.9 & 0.03 & \textbf{35.37} & 39.4 & 0.03 & \textbf{32.92} & 78.0 & 0.04 & \underline{29.64} & 132.2 & 0.05 & \textbf{27.77} \\
    \bottomrule
  \end{tabular}\\[2pt]
  {\footnotesize We evaluate factors up to $\times8$, given the spatial resolution of the test light fields.}
\end{table*}

\subsection{Experimental setup}
\label{sec:setup}

\paragraph{Datasets and metrics.}
We follow the BasicLFSR benchmark~\cite{ntire2023} on EPFL~\cite{epfl}, HCInew~\cite{hcinew}, HCIold~\cite{hciold}, INRIA~\cite{inria} and STFgantry~\cite{stfgantry}, using the split of~\cite{lfdfnet}, the central $5\times5$ views and bicubic degradation.
As in BasicLFSR, images are converted to YCbCr: the networks super-resolve the Y channel, Cb and Cr are upsampled bicubically, and PSNR is computed on Y and averaged over views, then scenes, then datasets.

\paragraph{Training.}
Following SEIIF~\cite{seiif}, we train one arbitrary-scale \method model per encoder with factors from 1.5 to 4 in steps of 0.5, so $\times2$--$\times4$ are in-scale and $\times6$ and $\times8$ out-of-scale factors. Each encoder outputs $C=64$ feature channels. Each model is evaluated across all tested factors. For the fixed-scale comparisons, we fine-tune at $\times2$ and $\times4$ as in SEIIF.
We train with an $\ell_1$ loss between $\hat{H}$ and the ground-truth Y channel of all $UV$ views.
The disparity head and the SH heads are initialized to zero, so training begins without parallax and with primitive colours equal to the sampled LR values. The shape head, the scale-conditioning branch and the residual branch also start at zero.
The opacity head is biased so that the initial opacities of the $K=9$ primitives of a macro pixel sum to about 0.8, with their centres initialized by the template lattice.
The initial output is therefore the normalized splat of the LR input.

\paragraph{Evaluation.}
\Cref{tab:fixed} compares eight encoders with their original upsamplers and with \method. Baseline results are from BasicLFSR~\cite{ntire2023} and, for RASLF, from~\cite{raslf}.
For \cref{tab:arbitrary}, we follow the SEIIF~\cite{seiif} protocol, combining DistgSSR and EPIT with SEIIF and with view-wise LIIF~\cite{liif}, LTE~\cite{lte}, DIIF~\cite{diif} and CiaoSR~\cite{ciaosr}. PSNR, FLOPs and parameter counts of these combinations and the baselines are from SEIIF, and those of SAEIIF from~\cite{saeiif}. We also apply GSASR~\cite{gsasr} view by view, instantiating the per-view Gaussian sets of \cref{fig:teaser}a, and train it with the same data and training strategy as \method. All these methods are timed on a single Tesla V100 with a $5\times5\times32\times32$ input.

\subsection{Comparison with existing upsamplers}
\label{sec:comparison}

\paragraph{Generality across encoders.}
\method improves the average PSNR of all eight encoders at both factors, with $\times4$ gains of 0.34--0.70\,dB and improvements on every dataset for every encoder (\cref{tab:fixed}).
These results show that \method can improve reconstruction even when the encoder already aggregates spatial and angular information.
For every encoder, the gain is larger at $\times4$ than at $\times2$.
STFgantry, which has the widest disparity range, receives the largest gain for most encoders, while gains on the narrow-baseline Lytro datasets are generally smaller at $\times2$.

\begin{figure}[t]
  \centering
  \begin{subfigure}{\linewidth}
    \centering
    \includegraphics[width=\linewidth,trim=0 2.7 0 10.6,clip]{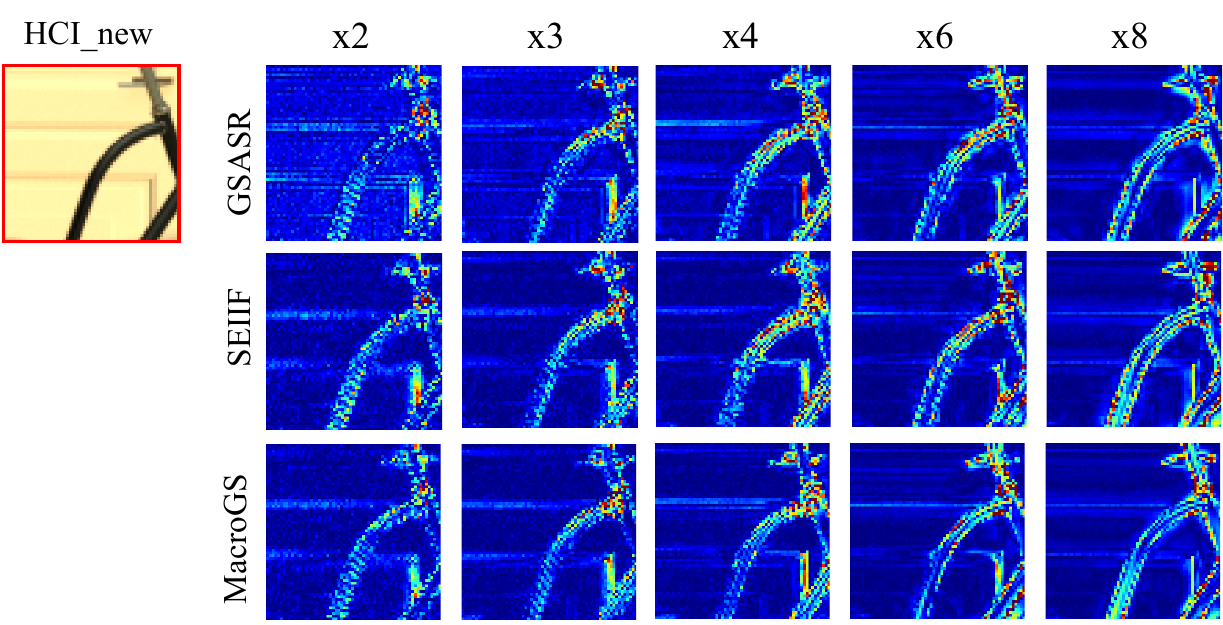}
    \caption{DistgSSR encoder}
    \label{fig:error_maps_distgssr}
  \end{subfigure}\\[4pt]
  \begin{subfigure}{\linewidth}
    \centering
    \includegraphics[width=\linewidth,trim=0 2.8 0 10.8,clip]{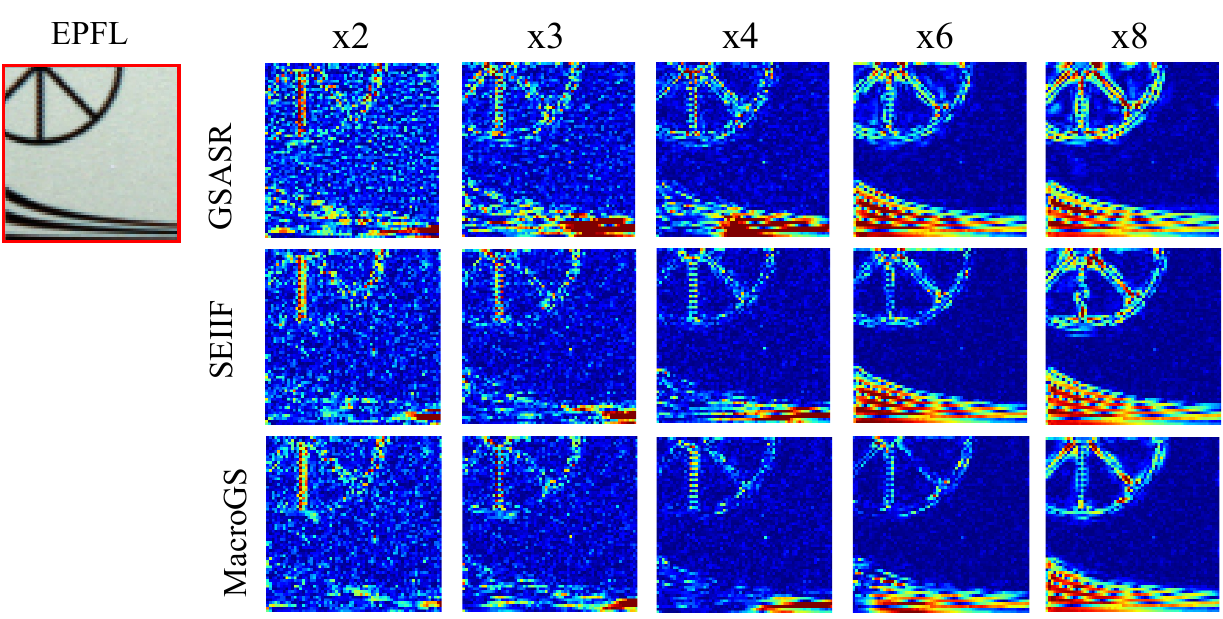}
    \caption{EPIT encoder}
    \label{fig:error_maps_epit}
  \end{subfigure}
  \caption{Error maps of GSASR, SEIIF and \method from $\times2$ to $\times8$, using (a) DistgSSR and (b) EPIT encoders.}
  \label{fig:error_maps}
\end{figure}

\begin{figure}[t]
  \centering
  \begin{subfigure}{\linewidth}
    \centering
    \includegraphics[width=\linewidth]{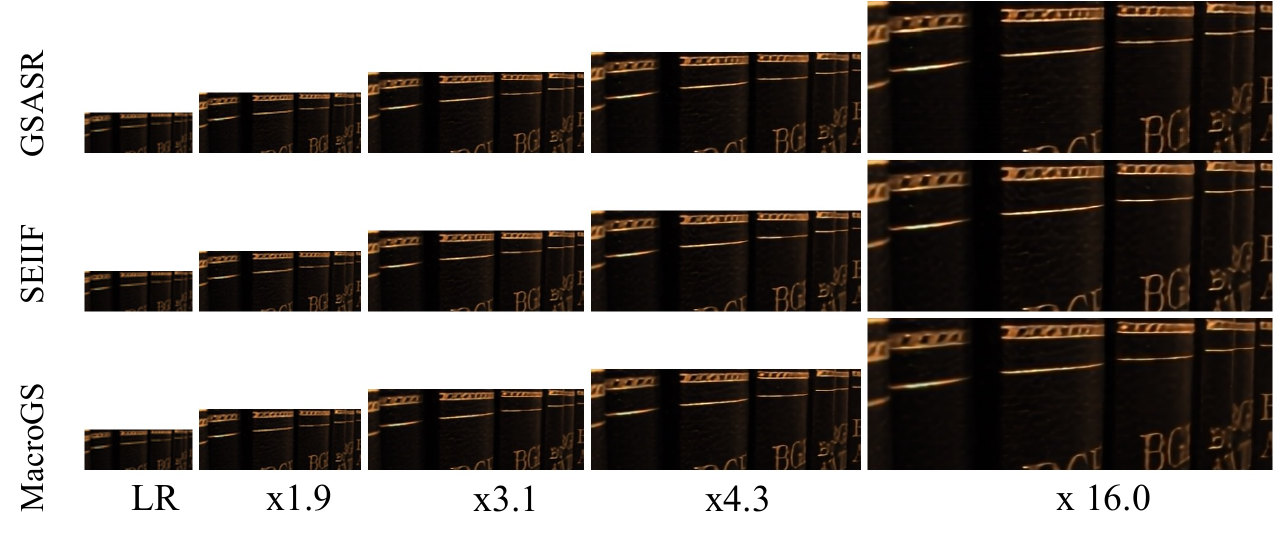}
    \caption{DistgSSR encoder}
    \label{fig:arbitrary_visual_distgssr}
  \end{subfigure}\\[4pt]
  \begin{subfigure}{\linewidth}
    \centering
    \includegraphics[width=\linewidth]{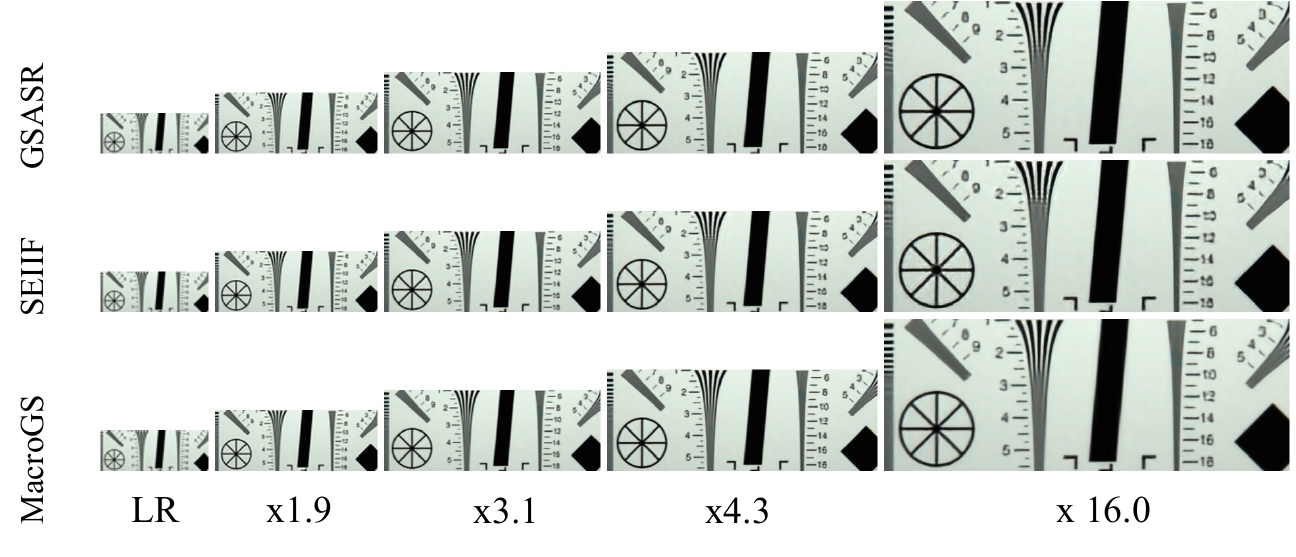}
    \caption{EPIT encoder}
    \label{fig:arbitrary_visual_epit}
  \end{subfigure}
  \caption{Arbitrary-scale results of GSASR, SEIIF and \method at $\times1.9$, $\times3.1$, $\times4.3$ and $\times16$, using (a) DistgSSR and (b) EPIT encoders.}
  \label{fig:arbitrary_visual}
\end{figure}

\paragraph{Arbitrary-scale comparison.}
Against GSASR, which predicts a separate primitive set for each view, \method is more accurate at every factor with both encoders while using far fewer parameters and decoding an order of magnitude faster (\cref{tab:arbitrary}); sharing one set across views thus comes with no loss in accuracy.
Among all compared decoders, \method is the most accurate in eight of the ten encoder--scale settings and close to the best in the other two, and at in-scale factors its advantage over SEIIF grows with the factor.
\method also decodes fastest at every factor, and its latency grows only slowly with the factor, whereas implicit decoders slow down as the number of queried output coordinates increases. At $\times8$ it is about three times as fast as SEIIF, although its FLOPs are higher, mainly because of the HR residual branch (\cref{sec:ablation}).
In \cref{fig:error_maps}, errors concentrate around object boundaries and grow with the factor. GSASR leaves the largest boundary errors and more scattered errors in flat regions, and \method reduces both, most visibly at the lower factors.
In \cref{fig:arbitrary_visual}, LFs first downsampled spatially by a factor of~4 are super-resolved at progressively larger factors that are not used in training. At $\times16$, \method keeps the converging lines of the resolution chart apart over a longer span than GSASR and SEIIF (\cref{fig:arbitrary_visual_epit}).

\suppressfloats[t]

\subsection{Learned disparity and scene geometry}
\label{sec:geometry}

\method learns disparity only through reconstruction, so we examine its correlation with scene geometry.
At $\times4$, we average the two components of each primitive disparity and take the mean over the $K$ primitives of a macro pixel, weighted by central-view opacity.
The reference disparity is the ground truth on HCIold and an SPO estimate~\cite{zhang2016spo} on HCInew.
Each macro pixel covers $4\times4$ HR pixels at this scale, so we average the reference over them.
We compare the two at textured locations, where the central-view gradient magnitude exceeds the scene median.
Pearson's $r$ measures linear correlation, invariant to positive rescaling and additive offsets; Spearman's $\rho$ measures rank correlation, indicating how well relative depth ordering is preserved.

\begin{figure}[t]
  \centering
  \captionof{table}{Correlation between opacity-weighted mean primitive disparities and scene references at $\times4$, on textured LR-grid locations.}
  \label{tab:disp_ref}
  {\footnotesize
  \setlength{\tabcolsep}{6pt}
  \begin{tabular}{lcccc}
    \toprule
    & \multicolumn{2}{c}{HCIold} & \multicolumn{2}{c}{HCInew} \\
    \cmidrule(lr){2-3}\cmidrule(lr){4-5}
    Encoder & $r\uparrow$ & $\rho\uparrow$ & $r\uparrow$ & $\rho\uparrow$ \\
    \midrule
    DistgSSR & 0.960 & 0.931 & 0.918 & 0.932 \\
    EPIT & 0.873 & 0.850 & 0.850 & 0.901 \\
    RASLF & \textbf{0.971} & \textbf{0.950} & \textbf{0.938} & \textbf{0.951} \\
    \bottomrule
  \end{tabular}
  }
  \par\vspace{10pt}
  {\footnotesize
  \setlength{\tabcolsep}{0pt}
  \setlength{\dispw}{\dimexpr\linewidth-11pt\relax}
  \begin{tabular}{@{}c@{\hspace{3pt}}c@{}}
    & \makebox[0.2\dispw]{EPFL}\makebox[0.2\dispw]{HCIold}\makebox[0.2\dispw]{HCInew}\makebox[0.2\dispw]{INRIA}\makebox[0.2\dispw]{STFgantry} \\[1pt]
    \rotatebox{90}{\makebox[0.196\dispw]{Disparity}\makebox[0.196\dispw]{Central view}} &
    \includegraphics[width=\dispw]{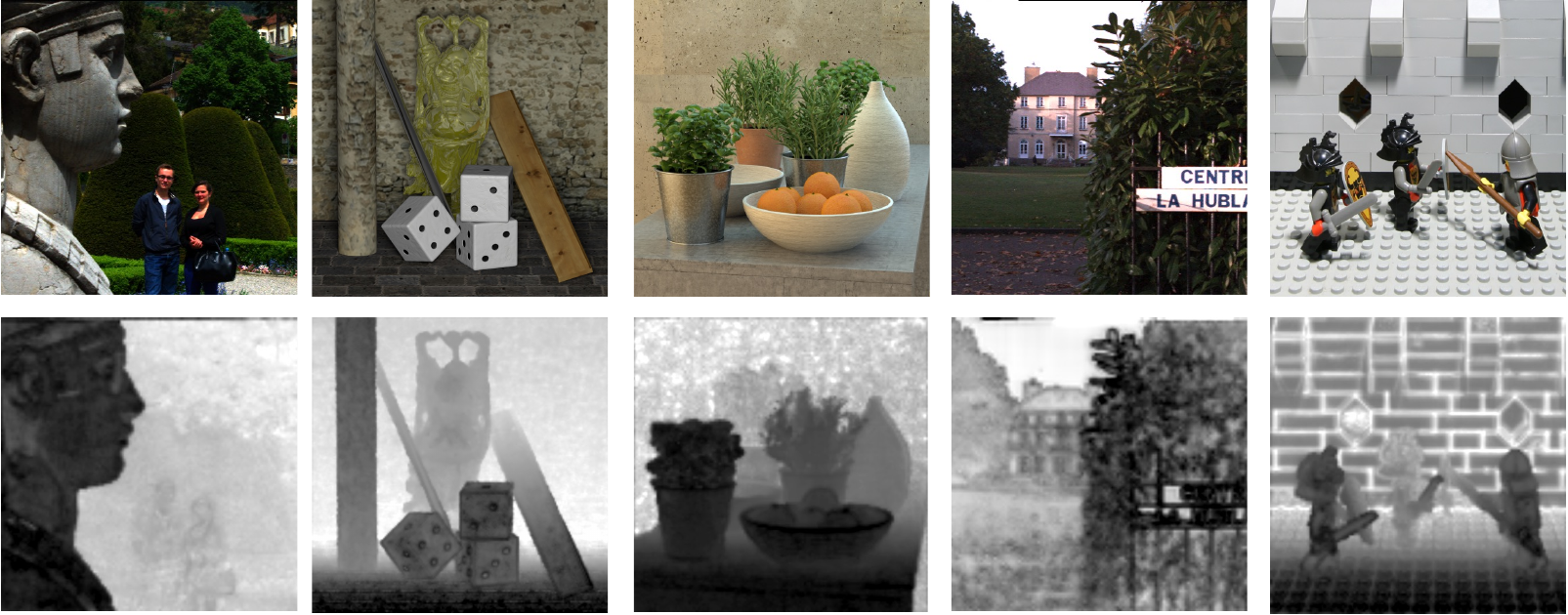} \\
  \end{tabular}
  }
  \captionof{figure}{Mean disparity maps from \method with RASLF at $\times4$, on one scene per benchmark. Each map is normalized separately, and darker is closer.}
  \label{fig:disp_maps}
\end{figure}

The mean disparity maps have $r$ and $\rho$ of at least 0.85 on both datasets, including HCIold with ground-truth disparity (\cref{tab:disp_ref}).
This correlation holds across convolutional (DistgSSR), Transformer-based (EPIT) and state-space (RASLF) encoders, with RASLF attaining the highest values on both datasets.
These strong correlations indicate that the learned disparities capture meaningful geometric structure, although they do not imply metrically accurate disparity, consistent with prior observations that geometry learned through reconstruction is optimized for the reconstruction objective~\cite{kalantari2016learning,chan2021understanding}.

\Cref{fig:disp_maps} shows mean disparity maps from the arbitrary-scale RASLF model without further optimization or supervision.
The maps distinguish foreground from background and follow slanted surfaces, such as the ground in the HCIold example.
The real-world examples also show foreground--background structure, although no reference disparity is available.
Texture-like variations remain on planar regions, such as the brick wall in STFgantry.

\subsection{Importance of the learned disparity}
\label{sec:disp_importance}

\begin{table}[t]
  \centering
  \caption{PSNR (dB) of disparity designs with RASLF at $\times2$, $\times4$ and $\times8$. 
  All variants retain primitive sets shared across views. 
  Macro-pixel and Primitive assign one disparity per macro pixel and per primitive, respectively.}
  \label{tab:disp_ablation}
  \footnotesize
  \setlength{\tabcolsep}{2.2pt}
  \begin{tabular}{llcccccc}
    \toprule
    Scale & Disparity & EPFL & HCInew & HCIold & INRIA & STFgantry & Avg. \\
    \midrule
    \multirow{3}{*}{$\times2$} & Zero & 35.16 & 38.28 & 45.37 & 36.83 & 41.14 & 39.36 \\
     & Macro-pixel & 35.19 & 38.39 & 45.46 & 36.92 & 41.52 & 39.50 \\
     & Primitive & \textbf{35.22} & \textbf{38.46} & \textbf{45.48} & \textbf{36.98} & \textbf{41.82} & \textbf{39.60} \\
    \midrule
    \multirow{3}{*}{$\times4$} & Zero & 30.06 & 31.87 & 38.10 & 32.10 & 32.64 & 32.95 \\
     & Macro-pixel & \textbf{30.12} & 31.96 & \textbf{38.16} & 32.11 & 32.81 & 33.03 \\
     & Primitive & 30.09 & \textbf{31.97} & \textbf{38.16} & \textbf{32.13} & \textbf{32.85} & \textbf{33.04} \\
    \midrule
    \multirow{3}{*}{$\times8$} & Zero & \textbf{25.40} & 27.62 & \textbf{32.81} & \textbf{27.33} & 26.23 & \textbf{27.88} \\
     & Macro-pixel & 25.38 & \textbf{27.64} & 32.60 & 27.31 & \textbf{26.36} & 27.86 \\
     & Primitive & 25.38 & 27.62 & 32.70 & 27.31 & 26.21 & 27.85 \\
    \bottomrule
  \end{tabular}
\end{table}

Explicit disparity pays off at in-scale factors, most where parallax is large (\cref{tab:disp_ablation}).
We retrain three RASLF-based variants that keep the shared primitive sets and fix the disparity to zero (Zero), share one disparity per macro pixel (Macro-pixel), or predict it per primitive (Primitive, the full model).
At $\times2$, each step improves every dataset, and per-primitive disparity helps most on STFgantry.
At $\times4$, both learned designs still improve on zero disparity but perform similarly, and at the out-of-scale factor $\times8$ the three designs are on par, so the benefit of per-primitive disparity is concentrated at $\times2$.
A plausible reason is that a disparity of $D$ HR pixels spans only $D/s$ LR pixels, so the displacements that aligned reading must resolve shrink as the factor grows.

\begin{table}[t]
  \centering
  \caption{Ablation with RASLF. Entries are average reconstruction PSNR (dB) / decoder FLOPs (G); parameters also count the decoder. Each variant changes the indicated component of the full model.}
  \label{tab:ablation}
  \footnotesize
  \setlength{\tabcolsep}{1.4pt}
  \begin{tabular}{lc*{3}{r@{\,/\,}l}}
    \toprule
    Variant & Params (M) & \multicolumn{2}{c}{$\times2$} & \multicolumn{2}{c}{$\times4$} & \multicolumn{2}{c}{$\times8$} \\
    \midrule
    \begin{tabular}[c]{@{}l@{}}Full \method\\(res, $K{=}9$, $R{=}3$)\end{tabular} & 0.37 & 39.60 & 16.2 & 33.04 & 39.4 & 27.85 & 132.2 \\
    \midrule
    w/o res & 0.29 & 39.22 & 8.5 & 32.97 & 8.7 & 27.88 & 9.6 \\
    $K{=}4$ & 0.37 & 39.42 & 11.6 & 32.92 & 34.7 & 27.78 & 127.1 \\
    $K{=}16$ & 0.37 & 39.62 & 22.6 & 33.05 & 46.0 & 27.86 & 139.5 \\
    $R{=}1$ & 0.21 & 39.47 & 10.9 & 32.95 & 34.2 & 27.67 & 127.0 \\
    $R{=}5$ & 0.53 & 39.64 & 21.4 & 33.06 & 44.6 & 27.88 & 137.5 \\
    \bottomrule
  \end{tabular}
\end{table}

\subsection{Ablation study}
\label{sec:ablation}

\Cref{tab:ablation} varies one component of \method at a time with RASLF.
The residual branch matters most at low factors: removing it lowers accuracy clearly at $\times2$, slightly at $\times4$ and not at $\times8$.
Without it, the decoding cost stays nearly constant across factors, because primitive prediction runs on the LR grid and splatting is cheap, whereas the branch runs at the output resolution and dominates the FLOPs at high factors.
The Gaussian part thus carries nearly all of the accuracy at high factors: at $\times8$, removing the branch cuts the FLOPs by an order of magnitude without loss of accuracy.
Accuracy rises from four to nine primitives per macro pixel and saturates at sixteen, so nine primitives suffice.
Iterative reading matters: a single round ($R=1$) lowers accuracy at every factor, most at the out-of-scale factor $\times8$, whereas two more rounds ($R=5$) improve it only marginally.
Three rounds thus offer a good trade-off between accuracy and cost.

%% file: sec/5_conclusion.tex
\section{Conclusion}
\label{sec:conclusion}

We presented \method, an arbitrary-scale LFSSR decoder that represents each macro pixel with a Gaussian set shared across views.
This design makes the primitive count independent of the number of views, while per-primitive disparities guide both iterative cross-view feature reading and view-conditioned rendering.
Learned solely through reconstruction, these disparities exhibit strong correlations with scene geometry.
\method improves all eight evaluated encoders over their original upsamplers, achieves the highest PSNR in eight of the ten arbitrary-scale settings, and provides a $10$--$18\times$ decoding speedup over our view-wise GSASR adaptation.
These results support shared, disparity-guided Gaussian primitives as an effective representation for accurate and efficient light field spatial super-resolution.
Future work will explore content-adaptive allocation of shared Gaussian primitives to better represent geometric boundaries and fine structures within a fixed primitive budget.